\documentclass[pdflatex,sn-mathphys]{sn-jnl}% Default with double column layout https://www.springer.com/journal/41060/
\usepackage{array}
\usepackage{graphicx}
\usepackage{amssymb}
\usepackage{xcolor}
\definecolor{revisor1}{HTML}{000000}%{E50000}%{873600} ROJO.
\definecolor{revisor2}{HTML}{000000}%{12840F}%{0B5345} VERDE.
\usepackage{hyperref} % Para el ORCID.
\usepackage{academicons} % Para el ORCID.
\usepackage{caption}
\usepackage{amsthm}
\usepackage[T1]{fontenc}
\usepackage{lmodern}

\usepackage{soul}

\usepackage{amsmath}
\usepackage{amsthm}
\usepackage{siunitx}

\usepackage{algorithm}
\usepackage{algpseudocode}
\algnewcommand{\LineComment}[1]{\State \(\triangleright\) #1}
\usepackage{makecell}

\jyear{2022}%

\theoremstyle{thmstyleone}%
\theoremstyle{thmstyletwo}%

\theoremstyle{thmstylethree}%
\newtheorem{definition}{Definition}%

\begin{document}

  \title{Feature Selection: A perspective on inter-attribute cooperation.}


\author*[1]{\fnm{Gustavo} \sur{Sosa-Cabrera} \href{https://orcid.org/0000-0002-9637-4319}{\includegraphics[scale=.5]{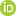}}} \email{gdsosa@pol.una.py} 

\author[1,2]{\fnm{Santiago} \sur{G\'{o}mez-Guerrero} \href{https://orcid.org/0000-0001-6363-0833}{\includegraphics[scale=.5]{orcid_16x16.png}}} \email{sgomez@pol.una.py}
\equalcont{These authors contributed equally to this work.}

\author[3]{\fnm{Miguel} \sur{Garc\'{i}a-Torres} \href{https://orcid.org/0000-0002-6867-7080}{\includegraphics[scale=.5]{orcid_16x16.png}}} \email{mgarciat@upo.es}
\equalcont{These authors contributed equally to this work.}

\author[1,2]{\fnm{Christian E.} \sur{Schaerer} \href{https://orcid.org/0000-0002-0587-7704}{\includegraphics[scale=.5]{orcid_16x16.png}}} \email{chris.schaerer@cima.org.py}
\equalcont{These authors contributed equally to this work.}

\affil*[1]{\orgdiv{Polytechnic School}, \orgname{National University of Asunci\'{o}n}, \orgaddress{\street{Mcal. L\'{o}pez}, \city{San Lorenzo}, \postcode{2111}, \state{Central}, \country{Paraguay}}}

\affil[2]{\orgdiv{CIMA}, \orgname{Center for Research in Mathematics}, \orgaddress{\street{C\'{e}sar L\'{o}pez Moreira}, \city{Asunci\'{o}n}, \postcode{1766}, \state{Capital}, \country{Paraguay}}}

\affil[3]{\orgdiv{Data Science and Big Data Lab}, \orgname{Universidad Pablo de Olavide}, \orgaddress{\street{Carretera de Utrera}, \city{Seville}, \postcode{41013}, \state{Provincia de Sevilla}, \country{Spain}}}

% Please provide an abstract of 150 to 250 words. The abstract should not contain any undefined abbreviations or unspecified references.

\abstract{High-dimensional datasets depict a challenge for learning tasks in data mining and machine learning. Feature selection is an effective technique in dealing with dimensionality reduction. It is often an essential data processing step prior to applying a learning algorithm. Over the decades, filter feature selection methods have evolved from simple univariate relevance ranking algorithms to more sophisticated relevance-redundancy trade-offs and to multivariate dependencies-based approaches in recent years. This tendency to capture multivariate dependence aims at obtaining unique information about the class from the intercooperation among features. This paper presents a comprehensive survey of the state-of-the-art work on filter feature selection methods assisted by feature intercooperation, and summarizes the contributions of different approaches found in the literature. Furthermore, current issues and challenges are introduced to identify promising future research and development.}

\keywords{Filter feature selection, Information-theoretic measures, High-order dependency, Feature intercooperation, Interaction, Synergy, Complementarity.}



\maketitle

%%%%%%%%%%%%%%%%%%%%%%%%%%%%%%%%%%%%%%%%%%
\section{Introduction}
%%%%%%%%%%%%%%%%%%%%%%%%%%%%%%%%%%%%%%%%%%
Large amounts of data are being generated in various fields of scientific research, including economic, financial, and marketing applications \cite{IM.chanda2009mining}. These data often have the characteristic of high dimensionality, which poses a high challenge for data analysis and knowledge discovery. Redundant and irrelevant features increase the learning difficulty of the prediction model, cause overfitting and reduce prediction performance \cite{YAO2022117002}. In order to use machine learning methods effectively, preprocessing of the data is essential. Feature selection has been proven effective in preprocessing high-dimensional data and in enhancing learning efficiency, from both theoretical and practical standpoints \cite{BLUM1997245,liu2012libro,JMLR:Guyon-2003}. Thus, to overcome problems arising from the high dimensionality of data, feature selection removes irrelevant and redundant dimensions by analyzing the entire dataset. 

Depending on whether the class label is used in the feature selection process or not, the feature selection methods can be categorized into supervised and unsupervised. Unsupervised feature selection is used to explore the dataset without the labeled data. The supervised feature selection uses the labels of samples to select the feature subset. In addition, supervised feature selection methods are usually grouped into three main categories: wrapper, embedded, and filter methods \cite{JMLR:Guyon-2003,liu2010feature,liu2012manipulating, Zhong2004UsingRS}. 

Wrappers search the space of feature subsets, using the classifier accuracy as the measure of utility for a candidate subset \cite{kohavi1997wrappers, wan2022r2ci}. The main advantage of such an approach is that the feature selection phase benefits from the direct feedback provided by the classifier. However, there are clear disadvantages in using the wrapper approach. The computational cost is huge, while the selected features are specific for the considered classifier. Embedded methods \cite{guyon2008feature} select features by determining which features are more important in the decisions of a predictive model. The wrapper and embedded methods can be categorized as classifier-dependent. On the other hand, strategies based on the filter approach can be categorized as classifier-independent \cite{macedo2019theoretical}.

The filter method approach evaluates the features' relevance based on the data's intrinsic properties, being independent of the learning process. In general, filters are relatively inexpensive in terms of computational efficiency; they are simple and fast, and, therefore, most of the designed methods pertain to this category \cite{bolon2016feature}. Furthermore, in real-world applications, many of the most frequently used feature selection algorithms are also filters \cite{liu2010feature}.

\textcolor{revisor2}{Recently, hybrid and ensemble methods were added to the general framework of feature selection in order to take advantage of both filter (computational efficiency) and wrapper (high performance) approaches \cite{almugren2019survey}.}

Over the decades, filter feature selection methods have evolved from simple univariate relevance ranking algorithms to more sophisticated relevance-redundancy trade-offs and to a multivariate dependencies-based approach in recent years. We refer to the latter as \textit{cooperativeness} (also known as complementariness \cite{RCDFS.chen2015feature}, synergy \cite{IWFS.zeng2015novel} and interaction \cite{CoRR:Jakulin_Bratko-2003}). 
 
Cooperating features are those that individually appear to be irrelevant or weakly relevant to the class; but taken in combination with other features, they are highly correlated to that target class.

\textcolor{revisor2}{The simplest example is probably the behavior of a XOR-patterned database of $3$ attributes $X_1$, $X_2$ and a class $C$. Since in this case $SU(X_1, C) = SU(X_2, C) =  0$, one is tempted to conclude that both $X_1$ and $X_2$ are irrelevant with respect to $C$. However, $MSU(X_1,X_2,C) > 0$; hence $X_1$ and $X_2$ intercooperate to determine the value of $C$. As a result a first simple rule for finding intercooperations can be ``find attributes $X$ such that $SU(X, C)$ equals $0$ or nearly $0$, then pair each of these with $C$ to check their relevance with respect to the class.''}

In particular, this relates to the fact that ignoring possible feature interdependencies results in subsets with redundancy and lack of cooperative features \cite{JMLR:Guyon-2003, jakulin2004testing}, which in turn cannot achieve optimal classification performance in most domains of interest \cite{xue2015survey}.

\textcolor{revisor1}{Finding relationships and dependencies among variables (that is, features and/or class) is usually accomplished by employing some measure. These relationships are relevance, redundancy, and cooperativeness (the latter being viewed as interaction, complementarity, or synergy). Generally, the filter methods are based on these concepts \cite{vergara2014review}.}

%A measure is used to find relationships and dependencies among variables (i.e. features and/or class). These relationships are relevancy, redundancy, and cooperativeness (interaction, complementarity or synergy). Generally, the filter methods are based on these concepts \cite{vergara2014review}. 

Several studies (see Section \ref{seccion:revision}) showed that taking into account high-order dependencies among variables can improve the performance of feature selection. \textcolor{revisor1}{More recently, Wan et al. \cite{wan2022r2ci} proposed a feature selection strategy using a filter-wrapper approach called $R2CI$, which takes into account multiple-feature correlations. In this paper \cite{wan2022r2ci}, the observations made on multiple dependencies are particularly interesting as they characterize complementarity and interaction, from the point of view of the two subsets of attributes (selected and non-selected) that are being generated into the search space. Undoubtedly, feature intercooperation has been drawing more attention in recent years.}
Thus, the literature on cooperativeness-based feature selection that considers feature dependence shows an increase despite early research on interaction information dating back to McGill (1954) \cite{mcgill1954multivariate} and subsequently advanced by Han (1980) \cite{han1980slepian}, Yeung (1991) \cite{yeung1991new}, Tsujishita (1995) \cite{tsujishita1995triple}, Guyon and Elisseeff (2003) \cite{JMLR:Guyon-2003}, Jakulin and Bratko (2004) \cite{jakulin2004testing} and Kojadinovic (2005) \cite{kojadinovic2005relevance}.

So, despite this research area receiving significant attention in recent years (most of the work has been published in the last decade), the problem is still challenging, and new algorithms emerge as alternatives to the existing ones.

In this paper, we focus on filter methods for feature selection based on feature intercooperation. We provide a comprehensive survey of the state-of-the-art work, and a discussion of the open issues and challenges for future work. For all the reviewed algorithms we provide the year of their first appearance in the scientific literature; the chronological perspective of feature cooperativeness evolution is presented in this manner.

We expect this survey to attract attention from researchers working on different feature intercooperation paradigms to investigate further effective and efficient approaches to addressing new challenges in feature selection.

This review paper is structured as follows. In the next section, we provide the basements of feature evaluation.  Filter methods foundations are introduced in Section \ref{seccion:teoria_filtros}. Then, we will review the literature on feature selection methods based on feature intercooperation in Section \ref{seccion:revision}, followed by a general discussion about issues and future challenges in Section \ref{seccion:casos}. Finally, we present our principal conclusions and future research lines.

%%%%%%%%%%%%%%%%%%%%%%%%%%%%%%%%%%%%%%%%%%
\section{Feature Evaluation} % Information-theoretic measures and type of attributes and Interaction  Discovery}
\label{seccion:teoria_informacion}
%%%%%%%%%%%%%%%%%%%%%%%%%%%%%%%%%%%%%%%%%%
Evaluation measure is a key part of feature importance criterion, which forms the basis of feature selection methods \cite{liu2012libro}. The feature selection objective is to find the relevant features (individually or in cooperation) and to discard redundant and irrelevant features in order to preserve the information contained in the whole set of input variables with respect to the target class.

\textcolor{revisor1}{The traditional correlation proposed by Pearson only computes the correlation between two numeric features. The ranked correlation measures by Spearman and separately by Kendall \cite{croux2010influence} do the same for two ordinal or ranked variables. But in addition to features that express quantity or order, there are also qualitative features in real life.
Qualitative features are more generic in the sense that every numeric attribute can be made qualitative by employing discretization methods \cite{lavangnananda2017study}. Ordinal features are already qualitative in their nature. In this work, we look at correlations between two or many qualitative features. At present, information theory methods are the ones that allow to compute correlations between two or more qualitative attributes, thus opening the doors to research in generalized feature selection techniques.}

We use information theory measures to quantify relevance, redundancy, and cooperativeness. Here, we show these concepts and basic definitions as follows. 

\subsection{Bivariate information measures}
\subsubsection{Mutual Information}
Mutual information (also called information gain (IG) \cite{MKP:Quinlan:1993} or two-way interaction \cite{jakulin2004testing}) measures the amount of stochastic dependency between variables, hence it can be used as a bivariate measure of correlation.

\begin{definition}
Consider a discrete random variable $X$, with possible values $\{x_1, \ldots, x_k\}$ and probability mass function $P(X)$, and suppose we draw a series of $X$ values. The entropy $H$ of the variable $X$ is a measure of the uncertainty in predicting the next value of $X$ and is given by
\begin{equation} 
H(X) := -\sum_{i} P(x_i)\log_{2}(P(x_i)).
\end{equation}
\end{definition}

The mutual information $I(X,Y)$ measures the reduction in uncertainty about the value of $X$ when the value of $Y$ is known, as expressed in the next definition. 

%\begin{equation}
%I(X;Y) := H(X) - H(X|Y) = H(Y) - H(Y|X) = H(X) + H(Y) - H(X,Y), 
%\end{equation}

\begin{definition} For discrete random variables $X$ and $Y$,  the mutual information $I(X,Y)$ is
\begin{equation}
\renewcommand{\arraystretch}{1.5}
\begin{array}{@{} r @{} >{{}} c <{{}} @{} l @{} l @{} l @{} }
  I (X;Y)    & :=  & H(X) - H(X\mid Y)\\
             &  =  & H(Y) - H(Y\mid X)\\
             &  =  & H(X) + H(Y) - H(X,Y),\\
\end{array}
\end{equation}
where $H(X,Y)$ is the extension of $H(X)$ using joint probabilities $P(x_i, y_j)$ in the definition of entropy.
\end{definition}

It can be shown that $I(X;Y) = 0$ when $X$ and $Y$ are statistically independent.

\subsubsection{Symmetrical Uncertainty}
Because the mutual information tends to be larger for variables with more labels, it is convenient to normalize its values using both entropies, originating the Symmetrical Uncertainty (SU) measure \cite{CUP:Press_et_al:1988} expressed as 
\begin{equation}
	SU(X,Y) := 2 \left[ \frac{I(X;Y)}{H(X) + H(Y)}\right].
\end{equation}
$SU$ restricts its values to the range [0,1].

\subsection{Multivariate information measures}
\subsubsection{Interaction Information}
Interaction information \cite{mcgill1954multivariate} among multiple variables can be understood as the amount of information shared or bound up in a set of $n$ random variables, but cannot be found within any subset of those $n$ variables. Then, the interaction information among three variables (3-way interaction information) is given by

\begin{equation}\label{iiDefinition}
\renewcommand{\arraystretch}{1.5}
\begin{array}{@{} r @{} >{{}} c <{{}} @{} l @{} l @{} l @{} }
  I (X;Y;Z)  & :=  & I(X;Y\mid Z) - I(X;Y)\\
             &  =  & I(X;Z\mid Y) - I(X;Z)\\
             &  =  & I(Z;Y\mid X) - I(Z;Y).\\
\end{array}
\end{equation}

Unlike mutual information, the interaction information can be negative, positive, or zero \cite{ICAP.jakulin2005machine}.

\subsubsection{Multivariate Symmetrical Uncertainty}
To quantify the dependency among more than two variables, the Multivariate Symmetrical Uncertainty (MSU) \cite{sosa2019multivariate} has been proposed as a generalization of the $SU$ according to the following expression.

\begin{equation}\label{msuDefinition}
MSU(X_{1:n}) := \frac{n}{n-1} \left[ 1 - \frac{H(X_{1:n})}{\sum_{i=1}^{n} H(X_i)} \right],
\end{equation}
where $H(X_{1:n})$ is the extension of $H(X)$ using the joint probabilities of variables $X_1, \ldots, X_n$. Like the Symmetrical Uncertainty, $MSU$ restricts its values to the range $[0, 1]$.

\subsection{\textcolor{revisor1}{Type of dependencies in data}}
%Let $F$, $S$, and $C$ denote the original feature set, the selected feature subset, and the target class, respectively. We review the following feature weighting definitions found in the literature.

\textcolor{revisor1}{Generally, for the evaluation of attributes, the feature selection process is based on the typing of dependencies between variables.
According to the behavior of one or several attributes (as possible predictors of the class), dependencies can be classified into univariate/multivariate relevance and univariate/multivariate redundancy.} 
\textcolor{revisor1}{This section aims to present a concise review of these notions. In particular, we will address a special case of multivariate relevance, which we have designated as intercooperativeness.}

Let $F$, $S$, and $C$ denote the original feature set, the selected feature subset, and the target class, respectively. Through univariate and multivariate measures based on information theory, we offer definitions of relevance, redundancy, and intercooperativeness. Also, we characterize a variable as relevant, redundant, or intercooperative for these sets.

%\textcolor{magenta}{This section aims to present a concise review of these notions. In particular, we will address a special case of multivariate relevance, which we have designated as intercooperativeness. To this end, consider $F$, $S$, and $C$ as the original set of features, the selected subset of features, and the target class, respectively. Hence, we introduce, using univariate and multivariate measures based on information theory, both the definition of relevance, redundancy, and intercooperativeness; as well as the characterization of a variable as relevant, redundant, or intercooperative.}

\subsubsection{Relevance}
In the literature, several works \cite{bell2000formalism,caruana1994useful,koller1996toward} have made an effort to classify the features according to their contribution to the meaning of the class concept. In this context, feature relevance has arisen as a measure of the amount of relevant information that a feature may contain about the class, where the level of individual relevance is defined either in terms of mutual information as $I(f_i;C)$, or $SU(f_i,C)$ using Symmetrical Uncertainty analogously.

%\st{In this context, a feature is considered irrelevant if it contains no information about the class and therefore it is not necessary at all for the predictive task.}

In this context, a feature is considered irrelevant if it contains no information about the class and is unnecessary for the predictive task.

\subsubsection{Redundancy}
%Redundancy is generally defined in terms of feature correlation and thereby, it is quantifiable with the level of dependency among two or more features. In terms of information measures, a bivariate approach for feature redundancy is defined as $I(f_i;f_j)$, while a negative value of $I(f_i;S;C)$ indicates partial or complete redundancy in multivariate approach \cite{FRFS.wang2013selecting, yu2004efficient, sosa2019multivariate}. Similarly, to measure the common portion of information received from a set of features, Multivariate Symmetrical Uncertainty can be used as $MSU(f_{1:n})$ where $f_{1:n}$ represents the feature set $f_1,...,f_n$. 

Redundancy is generally defined in terms of feature correlation, and thereby, it is quantifiable with the level of dependency among two or more features. In terms of information measures, a bivariate approach for feature redundancy is defined as $I(f_i;f_j)$, while a negative value of $I(f_i;S;C)$ indicates partial or complete redundancy in multivariate approach \cite{FRFS.wang2013selecting, yu2004efficient, sosa2019multivariate}. Similarly, to measure the common portion of information received from a set of features, Multivariate Symmetrical Uncertainty can be used as $MSU(f_{1:n})$ where $f_{1:n}$ represents the feature set $f_1, \ldots ,f_n$.

\subsubsection{Intercooperativeness}%/Interaction/Complementarity/Synergy}
Intuitively, the intercooperativeness measures the amount of information received from grouped features, instead of separate features  \cite{sosa2019multivariate,vergara2014review,ICAP.jakulin2005machine}. This concept, in which a set of features cooperate to predict the class label, can be quantified according to a positive value of the expression $I(f_1; f_2; \ldots;f_n; C)$. Similarly, $MSU(f_1, f_2, \ldots, f_n, C)$ can be used as a measure of the cooperative association of two o more features $f_1, f_2, \ldots , f_n$ along with the $C$ class variable.

Feature cooperativeness must be measured with the target variable, that is, compute the relevance of a feature to the class with at least another feature presented. Therefore, three-way dependency is the minimum order for the evaluation method of intercooperativeness.

%\textcolor{magenta}{minimal order o minimum order?}

Note that to describe this same concept, the terms interaction, synergy, and complementarity are used interchangeably throughout the literature. However, in Section \ref{seccion:casos}, we shall argue for a more precise interpretation for each case.

%\textcolor{magenta}{Hasta aqui}

%%%%%%%%%%%%%%%%%%%%%%%%%%%%%%%%%%%%%%%%%%%%%%%%%%%%%%%%%%%%%%%%%%%%%%%%%%%%%%%%%%%%%%%%%%%%%%%%%%%%%%%%%%%%%%%%%%%%%%%%%%%%%%%%%%%%%%%%%%%%%%%%%%%%%%%%%%%%%%%%%%%%%

\section{Filter Methods}
\label{seccion:teoria_filtros}
%%%%%%%%%%%%%%%%%%%%%%%%%%%%%%%%%%%%%%%%%%

A filter feature selection process is independent of any learning algorithm and relies on underlying attributes of data. Thereby, to evaluate the utility of features, a filter model depends on statistical criteria applied to data such as distance, dependency, information, consistency, and correlation \cite{ullah2017dimensionality}.

A filter feature selection method attempts to select the minimally sized subset of features according to a loop of subset generation (by search strategy) and its evaluation (by measure) until some stopping criterion is satisfied \cite{cai2018feature}. Based on these basic steps, an abstract algorithm for feature selection that shows the behavior of any filter method in a unified form is depicted in Algorithm \ref{alg:filtro}.

\begin{algorithm}[h]
\caption{A generalized filter method}\label{alg:filtro}
\begin{algorithmic}[1]
\Function{filter}{$F, S_0, \delta, S_{best}$} 
\LineComment{$F$: full feature set}
\LineComment{$S_{0}$: a subset from which to start the search}
\LineComment{$\delta$: a stopping criterion}
\LineComment{$S_{best}$: an optimal subset}
\State $S_{best} \gets S_{0}$ \Comment{initialize $S_{best}$}
\State $\gamma_{best} \gets evaluate(S_0,F,M)$ \Comment{evaluate $S_0$ by measure $M$}
\Repeat
\State $S \gets search\ strategy(F,S_{best})$ \Comment{generate next candidate subset}
\State $\gamma \gets evaluate(S,F,M)$ \Comment{evaluate $S$ by measure $M$}
\If {$\gamma$ is better than $\gamma_{best}$}
\State $\gamma_{best} \gets \gamma$ \Comment{update $\gamma_{best}$}
\State $S_{best} \gets S$ \Comment{update $S_{best}$}
\EndIf
\Until{$\delta$ \textbf{is} reached}
\State \textbf{return} $S_{best}$
\EndFunction
\end{algorithmic}
\end{algorithm}

\section{Feature-Intercooperation-based \\
filter methods}
\label{seccion:revision}
%%%%%%%%%%%%%%%%%%%%%%%%%%%%%%%%%%%%%%%%%%%%%%%%%%%%%%%
In the feature selection field, the detection and significance of higher order interactions between variables have been a matter of discussion and experimentation, especially in recent years.

%\st{This section briefly summarizes, to the best of our knowledge, the existing filter feature selection methods assisted by feature intercooperation from three aspects, which are the estimation of high-dimensional dependencies, the search techniques, and the number of higher-order interactions.}

In this section, we briefly summarize the existing filter feature selection methods assisted by feature intercooperation, looking at three aspects: the estimation of high-dimensional dependencies, the search techniques, and the number of higher-order interactions.

%%%%%%%%%%%%%%%%%%%%%%%%%%%%%%%%%%%%%%%%%%%%%%%%%%%%%%%%%
\subsection{\textcolor{revisor1}{Estimation of high-order interactions.}}
%%%%%%%%%%%%%%%%%%%%%%%%%%%%%%%%%%%%%%%%%%%%%%%%%%%%%%%%%
Information theoretic quantities, such as mutual information and its generalizations, have several advantages as measures of multiple variable dependence. They are inherently model-free and non-parametric, and exhibit only modest sensitivity to undersampling \cite{tit:mcgill:1954,CoRR:Jakulin_Bratko-2003}.
However, it has long been recognized that information theory measures, and many others, generally cannot be computed analytically for all possible subsets of dependent variables. 
As such, researchers have developed methods that can calculate the presence of nonlinear and high-dimensional dependencies efficiently and reasonably.

%%%%%%%%%%%%%%%%% FJMI %%%%%%%%%%%%%%%%%%%%%%%%%
Thus, instead of directly calculating the five-way interaction terms, which are computationally expensive, FJMI \cite{FJMI.TANG2019207} took into account two- through five-way interactions between features and the class variable to capture interactions. The approach is based on the fact that five-dimensional joint mutual information can be decomposed into a sum of two- through five-way interactions, which is easier to compute.

%%%%%%%%%%%%%%%%% CMICOT %%%%%%%%%%%%%%%%%%%%%%%%%
Shishkin et al. \cite{CMICOT.shishkin2016efficient} proposed the CMICOT method, which uses conditional mutual information (CMI) to identify joint interactions between multiple features (more than three). The technique is based on a two-stage greedy search for the approximate solution of high-dimensional CMI and binary representation of features that reduce the dimension of the space of joint distributions, to mitigate the effect of the sample complexity.

%%%%%%%%%%%%%%%%% RelaxMRMR %%%%%%%%%%%%%%%%%%%%%%%%%
Vinh et al. \cite{RelaxMRMR.vinh2016can} proposed a higher dimensional MI-based feature selection method called RelaxMRMR. To capture higher-order feature interactions, the authors identified the assumptions that can be relaxed for decomposing the full joint mutual information criterion into lower-dimensional MI quantities.

%%%%%%%%%%%%%%%%% IWFS %%%%%%%%%%%%%%%%%%%%%%%%%
To explicitly treat feature interaction, Zeng et al. \cite{IWFS.zeng2015novel} proposed a complementarity-based ranking method called IWFS. The approach is based on interaction weight factors, a variation of three-way interaction that can measure redundancy and complementarity between features.

%%%%%%%%%%%%%%%%% CMIFS %%%%%%%%%%%%%%%%%%%%%%%%%
Based on the link between interaction information and conditional mutual information, Cheng et al. \cite{CMIFS.cheng2011conditional} proposed a greedy algorithm called CMIFS, which considers not only the competition among features but also the cooperation. This criterion takes account of both redundancy and synergy interactions of features and identifies discriminative features.

%%%%%%%%%%%%%%%%% IGFS %%%%%%%%%%%%%%%%%%%%%%%%%
El Akadi et al. \cite{IGFS.el2008powerful} proposed an evaluation function called IGFS. It takes into account different features interaction without increasing the computational complexity, and is based on the individual Mutual Information and a compromise (made by the mean of Interaction Gain) between features redundancy and features interaction. 

%%%%%%%%%%%%%%%%% OFS-MI %%%%%%%%%%%%%%%%%%%%%%%%%
Chow and Huang \cite{OFS-MI.chow2005estimating} combined a pruned Parzen window estimator and the quadratic mutual information for the effective and efficient estimation of high-dimensional mutual information. 

With this contribution, Chow and Huang developed a feature selection method called OSF-MI which can identify the salient features and analytically estimate the appropriate feature subsets.

%%%%%%%%%%%%%%%%%%%%%%%%%%%%%%%
\subsection{Search Techniques.}
%%%%%%%%%%%%%%%%%%%%%%%%%%%%%%%
%\st{Feature selection can be viewed as a search problem, with each state in the search space specifying a subset of the relevant features. An exhaustive method can be used for this purpose in theory, but is quite impractical and fact there are very few feature selection methods that use an exhaustive search} \cite{xue2015survey}.
%\st{Therefore, heuristic search strategies, such as greedy search, best-first search, and genetic-algorithmic search can be used in a backward elimination or a forward selection process for obtaining possible features as a suboptimal solution.
%However, feature selection problems have a large search space, which is very complex due to feature interaction. To overcome such issues, filter methods that can restrict the solution search space and makes the computation more tractable have become essential.}

Feature selection can be viewed as a search problem, with each state specifying a subset of the relevant features in the search space. An exhaustive method can be used for this purpose in theory but is quite impractical, and in fact, very few feature selection methods use an exhaustive search \cite{xue2015survey}. Therefore, heuristic search strategies such as greedy, best-first, and genetic-algorithmic, can be used in a backward elimination or forward selection process for obtaining possible features as a suboptimal solution.
However, feature selection problems have a large search space, which is very complex due to feature interaction. To overcome such issues, filter methods that can restrict the solution search space and make the computation more tractable have become essential.

%%%%%%%%%%%%%%%%% SAFE %%%%%%%%%%%%%%%%%%%%%%%%%
Recently, Singha and Shenoy \cite{SAFE.singha2018adaptive} proposed an adaptive method called SAFE which uses an adaptive 3-way cost function that uses redundancy–complementarity ratio to automatically update the trade-off rule between relevance, redundancy, and complementarity. This approach uses the best-first search strategy, which offers the best compromise solution.

%%%%%%%%%%%%%%%%%% IMFS-FD %%%%%%%%%%%%%%%%%%%%%%%%
Since it is necessary to balance accuracy and complexity in high-order interactions, Tang et al. \cite{IMFS-FD.tang2018interaction} presented a method called IMFS-FD to obtain a set of features that preserves $k$-way important interactions but does not intend to interpret all possible interactions reducing the search space.

%%%%%%%%%%%%%%%%% FGMMI %%%%%%%%%%%%%%%%%%%%%%%%%
Mohammadi et al. \cite{FGMMI.mohammadi2017multivariate} implemented the feature grouping based on multivariate mutual information (FGMMI), which discovers hidden relations between more than two features at the same time. This method aims to construct groups by using the $k$-means algorithm on a computed MI matrix which divides data into clusters and finally computes MMI for all of the features in each group to select each group's feature having the maximum relevance.

%%%%%%%%%%%%%%%%% RJMIM %%%%%%%%%%%%%%%%%%%%%%%%%
Peng and Liu \cite{RJMIM.peng2016rjmim} proposed the RJMIM method that employs a forward greedy search strategy to find and select the features with high discriminative power by measuring both the joint mutual information and the interaction information between the features already selected and candidate features.

%%%%%%%%%%%%%%%%% NIWFS %%%%%%%%%%%%%%%%%%%%%%%%%
Zeng et al. \cite{NIWFS.zeng2015mixed} proposed a feature ranking algorithm called NIWFS. It is based on neighborhood rough sets that can be used to search for interacting features. Since redundant features produce negative influence and interaction features produce positive influence in predicting, this approach first computes the neighborhood mutual information between a feature and the target and then adjusts it by manipulating the interaction weight factor, which can reflect the information of whether a feature is redundant or interactive.

%%%%%%%%%%%%%%%%% FIM %%%%%%%%%%%%%%%%%%%%%%%%%
%\textcolor{magenta}{Bennasar et al. \cite{FIM.bennasar2013feature} employs feature interaction - a maximum of the minimum criteria to select the feature with the strongest relevance to the class label and the highest minimum interaction with the already selected. The FIM method is based on three-way interaction information using a forward greedy search algorithm to select relevant and non-redundant features.}

Bennasar et al. \cite{FIM.bennasar2013feature} employs feature interaction -- a maximum of the minimum criteria to select the feature that has the strongest relevance to the class label and the highest minimum interaction with the already selected. This method called FIM, is based on three-way interaction information using a forward greedy search algorithm to select relevant and non-redundant features.

%%%%%%%%%%%%%%%%% BIFS %%%%%%%%%%%%%%%%%%%%%%%%%
To identify all possible feature interactions of maximum size, Sui \cite{BIFS.sui2013information} proposed a BIFS method which is constructed by two main processes: forward identification to identify binary interactions and backward selection where irrelevant feature interaction subsets will be deleted from subsets ranked based on information gain per feature (IGFS).

%%%%%%%%%%%%%%%%% DSplusMII %%%%%%%%%%%%%%%%%%%%%%%%%
Zhang and Hancock~\cite{DSplusMII.zhang2011graph} presented a method called DSplusMII, which utilizes the multidimensional interaction information criterion and dominant sets for feature selection. This approach can consider third or higher-order feature interactions and limits the resulting search space using dominant set clustering, which separates features into clusters in advance.

%%%%%%%%%%%%%%%%% INTERACT %%%%%%%%%%%%%%%%%%%%%%%%%
Zhao and Liu~ \cite{INTERACT.zhao2009searching} proposed the INTERACT method, which finds interacting features based on a feature sorting metric using data consistency. Contrary to an evaluation based on mutual information, the inconsistency measure is monotonic, allowing an efficient search to explore feature interactions.

%Zhao and Liu~ \cite{INTERACT.zhao2009searching} proposed the INTERACT method, which finds interacting features based on a feature sorting metric using data consistency. Contrary to an evaluation based on mutual information, the inconsistency measure is monotonic and hence allows an efficient search to explore feature interactions.

%%%%%%%%%%%%%%%%% DISR %%%%%%%%%%%%%%%%%%%%%%%%%
For a complementary attribute of an already selected attribute to have a 
%In order to a complementary attribute of one already selected has a 
much greater probability of being selected, 
Meyer and Bontempi~\cite{DISR.meyer2006use} have introduced a method called DISR. Its goal function uses symmetrical relevance and considers the net effect of redundancy and complementarity in the search process. They show that a set of attributes can return information on the class variable that is higher than the sum of the informations from each attribute taken individually.

%%%%%%%%%%%%%%%%%%%%%%%%%%%%%%%%%%%%%%%%%%%%
\subsection{Number of higher-order interactions.} % number of higher-order interactions
%%%%%%%%%%%%%%%%%%%%%%%%%%%%%%%%%%%%%%%%%%%%
Despite being hard to measure directly, the interaction and the candidate interactions grow exponentially with the number of features (i.e., the number of variables when considering interactions increases by several orders of magnitude), and higher-order interactions have enormous potential for improving the performance of feature selection. This illustrates why the exploration of high-order interactions is a challenge where increasingly efficient methods have been developed to take into account both 3-way, 4-way and 5-way interactions and can possibly extended to the case of full higher-order terms.

%Despite being hard to directly measure the interaction and the candidate interactions grows exponentially with the number of features (i.e., the number of variables when considering interactions increase by several orders of magnitude), higher-order interactions have enormous potential for improving the performance of feature selection. This illustrates why the exploration of high-order interactions is a challenge where increasingly efficient methods have been developed to take into account both 3-way, 4-way and 5-way interactions and can possibly extended to the case of full higher-order terms.

%%%%%%%%%%%%%%%%% FS-RRC %%%%%%%%%%%%%%%%%%%%%%%
Recently, for instance, Wang et al.\cite{wang2021.MRMI} proposed an algorithm called MRMI to explore three-way interactions. Future works include how it can be extended to the case of higher order terms to select strongly relevant and possibly more interactive features. 

%%%%%%%%%%%%%%%%% FS-RRC %%%%%%%%%%%%%%%%%%%%%%%
To retain the features with the greatest complementarity in the selected feature subset during the progress of feature selection, Li et al.\cite{li2020.FSRRC} proposed a new algorithm, FS-RRC, which computes the complementarity score of two features and the class (three-way interactions).

%To retain the features with the greatest complementarity with the selected feature subset in the progress of feature selection, Li et al.\cite{li2020.FSRRC} proposed a new algorithm called FS-RRC which computes the complementarity score of two features and the class (three-way interactions).

%%%%%%%%%%%%%%%%% IIFS %%%%%%%%%%%%%%%%%%%%%%%%%
Pawluk et al. \cite{IIFS.pawluk_teisseyre_mielniczuk_2019} proposed a feature selection method named IIFS that considers both 3-way and 4-way interactions. Based on interaction information, they prove some theoretical properties of the novel criterion and the possibility that it may be extended to the case of higher-order terms.

%%%%%%%%%%%%%%%%% RCDFS %%%%%%%%%%%%%%%%%%%%%%%%%
Since the dependence among features is related to both redundancy and complementariness, Chen et al. \cite{RCDFS.chen2015feature} proposed a method called RCDFS where the complementary correlation of features is explicitly separated from redundancy. In this approach, a modification item concerning feature complementariness is introduced in the evaluation criterion in order to identify interaction among more than two features.

%%%%%%%%%%%%%%%%% GlobalFS %%%%%%%%%%%%%%%%%%%%%%%%%
Vinh et al. \cite{GlobalFS.vinh2014reconsidering} introduced GlobalFS, which can automatically select the number of features to be included and can assess high-order feature dependency via high dimensional mutual information. However, it is only suitable for problems with a small to medium number of features, e.g., several tens. 

%%%%%%%%%%%%%%%%% FRFS %%%%%%%%%%%%%%%%%%%%%%%%%

Wang et al. \cite{FRFS.wang2013selecting} proposed a rule-based feature selection algorithm FRFS for not only identifying and removing irrelevant and redundant features, but also preserving the interactive ones. The method employs the FOIL algorithm with a restriction to generate classification rules to collect the features whose values appear in the antecedents of the rules generated. Then, it eliminates irrelevant and redundant features while considering multi-way feature interactions.

%Wang et al. \cite{FRFS.wang2013selecting} proposed a rule-based feature selection algorithm FRFS for not only identifying and removing irrelevant and redundant features but also preserving interactives ones. The method employs the FOIL algorithm with a restriction to generate classification rules in order to collect the features whose values appeared in the antecedents of the rules generated. Then, it eliminates irrelevant and redundant features while considering the multi-way feature interactions.

%%%%%%%%%%%%%%%%%% mIMR %%%%%%%%%%%%%%%%%%%%%%%%
Bontempi and Meyer \cite{mIMR.bontempi2010causal} presented mIMR, a causal filter criterion based on three-way interaction that aims to select a feature subset where the most informative variables are the ones having both high mutual information with the class and high complementarity with the others.

%%%%%%%%%%%%%%%%% CMIM-2 %%%%%%%%%%%%%%%%%%%%%%%%%
To detect pairs of relevant variables that act complementarily in predicting the class, Vergara et al.\cite{CMIM-2.vergara2010cmim} proposed CMIM-2 as an improvement of the CMIM criterion. It maintains the advantages of the original criterion, but it solves the problem of variables that are relevant in pairs, changing the minimum function to the average function.  

%%%%%%%%%%%%%%%%%% IM %%%%%%%%%%%%%%%%%%%%%%%%
Chanda et al. \cite{IM.chanda2009mining} proposed an Interaction Mining (IM) approach to capture the multivariate inter-dependencies (synergy and redundancy) among features, so they employ this $k$-way interaction information to improve a feature subset selection that has significant interactions with the class variable.

%Chanda et al. \cite{IM.chanda2009mining} proposed an Interaction Mining (IM) approach to capture the multivariate inter-dependencies (synergy and redundancy) among features, so they employ this $k$-way interaction information to improve feature subset selection that have significant interactions with the class variable.

%%%%%%%%%%%%%%%%% ICAP/IC %%%%%%%%%%%%%%%%%%%%%%%%%
Jakulin and Bratko \cite{ICAP.jakulin2005machine} introduced interaction information to measure feature interactions and proposed a feature selection method called ICAP which can detect two-way (one feature and the class) and three-way (two features and the class) interactions.

%\textcolor{magenta}{}
%%%%%%%%%%%%%%%%%%%%%%%%%%%%%%%%%%%%%%%%%%
%\section{Feature-Intercooperation-based filter methods}
%\label{seccion:revision}
%To the best of our knowledge, the following are the existing filter feature selection methods assisted by feature intercooperation, ordered by year of publication.
%\begin{itemize}[leftmargin=3.5em, noitemsep]
%\end{itemize}
%%%%%%%%%%%%%%%%%%%%%%%%%%%%%%%%%%%%%%%%%%

To close this section we would like to mention that using a chronological perspective, we observe how this research topic receives greater attention from researchers since 2005. It can be concluded that Jakulin's work has had a significant influence on the development of methods for feature selection based on higher-order interaction. 
\begin{figure*}[t]
\centering
\includegraphics[scale=0.75]{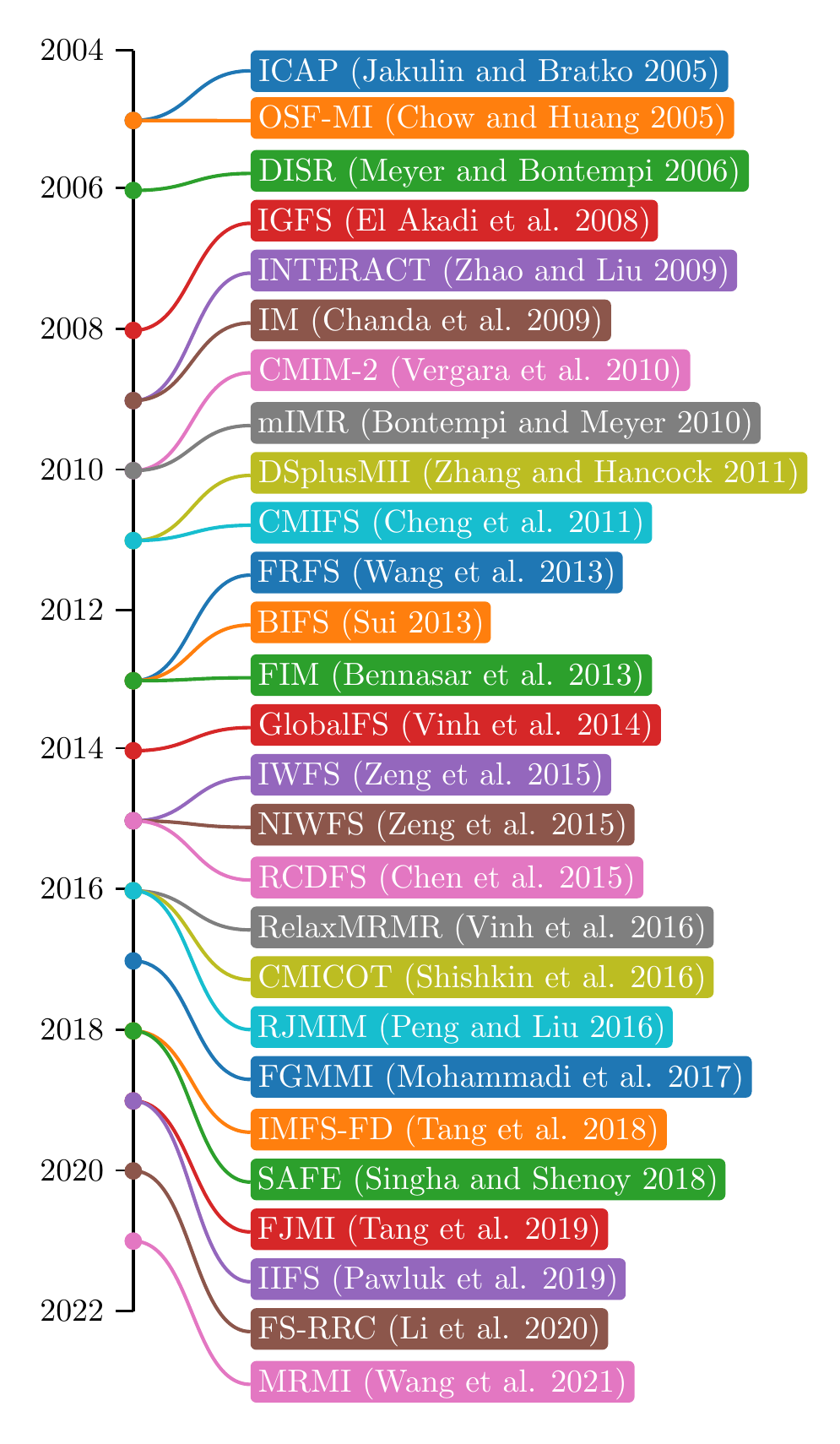}
\caption{Timeline of publications for filter methods based on feature intercooperation. Note that publications in this field are not numerous and represent the results of research initiated in 2005.}
\end{figure*}   

In addition, the entire list of $27$ algorithms surveyed, sorted by name, including also full name and reference, is shown in Table \ref{tab:algoritmos}.

\begin{table*}[t]
\centering 
\protect\caption{Filter Methods based on Feature Intercooperation sorted by name.
}
\scriptsize
\begin{tabular}{ccc}
\hline\hline\\  [-2ex]
\textbf{Method}	& \textbf{Full Name} & \textbf{Reference}\\
\hline\\[-1ex]
BIFS & \makecell{Binary Interaction based\\Feature Selection} & \cite{BIFS.sui2013information}\\
%CIFE & Conditional Informative Feature Extraction & \cite{CIFE.10.1007/11744023_6}\\
CMICOT & \makecell{Conditional Mutual Information with \\ Complementary and Opposing Teams} & \cite{CMICOT.shishkin2016efficient}\\
CMIFS & \makecell{Conditional Mutual Information\\Feature Selection} & \cite{CMIFS.cheng2011conditional} \\
CMIM-2 & \makecell{Conditional Mutual Information\\Maximization Version 2} & \cite{CMIM-2.vergara2010cmim}\\
%DFL & Discrete Function Learning & \citep{DFL.zheng2011feature}\\
DISR & \makecell{Double Input Symmetrical Relevance} & \cite{DISR.meyer2006use}\\
DSplusMII & DSplusMII & \cite{DSplusMII.zhang2011graph}\\
FGMMI & \makecell{Feature Grouping based on\\Multivariate Mutual Information} & \cite{FGMMI.mohammadi2017multivariate}\\
FIM & \makecell{Feature Interaction Maximisation} & \cite{FIM.bennasar2013feature}\\
FJMI & \makecell{Five-way Joint Mutual Information} & \cite{FJMI.TANG2019207}\\
FRFS & \makecell{FOIL Rule based\\Feature Subset Selection} & \cite{FRFS.wang2013selecting}\\
FS-RRC & \makecell{Feature Selection based on relevance,\\ redundancy and complementarity} & \cite{li2020.FSRRC} \\
GlobalFS & \makecell{Global Feature Selection} & \cite{GlobalFS.vinh2014reconsidering}\\ 
ICAP/IC & \makecell{Interaction Capture} & \cite{ICAP.jakulin2005machine}\\
IGFS & \makecell{Interaction Gain for Feature Selection} & \cite{IGFS.el2008powerful}\\
IIFS & \makecell{Interaction Information\\Feature Selection} & \cite{IIFS.pawluk_teisseyre_mielniczuk_2019}\\
IM  & \makecell{Interaction Mining} & \cite{IM.chanda2009mining}\\
IMFS-FD & \makecell{Interaction-based Feature Selection\\using Factorial Design} & \cite{IMFS-FD.tang2018interaction}\\
INTERACT & INTERACT & \cite{INTERACT.zhao2009searching}\\
IWFS & \makecell{Interaction Weight based\\Feature Selection} & \cite{IWFS.zeng2015novel}\\
%JMI & Joint Mutual Information & \cite{JMI.yang1999feature}\\
%JMIM & Joint Mutual Information Maximisation & \cite{JMIM.bennasar2015feature}\\
mIMR & min-Interaction Max-Relevance & \cite{mIMR.bontempi2010causal}\\
MRMI & Max-Relevance Max-Interaction & \cite{wang2021.MRMI}\\
%MMIMRSC & Maintaining Mutual Information and Minimizing Redundancy-Synergy Coefficient & \cite{MMIMRSC.yang2004feature}\\
%MRMC & Maximum Relevancy Maximum Complementary & \cite{MRMC.chernbumroong2015maximum}\\
NIWFS & \makecell{Neighborhood Interaction Weight\\based Feature Selection} & \cite{NIWFS.zeng2015mixed}\\
OFS-MI & \makecell{Optimal Feature Selection using\\Mutual Information} & \cite{OFS-MI.chow2005estimating}\\
RCDFS & \makecell{Redundancy-Complementariness\\Dispersion Feature Selection} & \cite{RCDFS.chen2015feature}\\
RelaxMRMR & RelaxMRMR & \cite{RelaxMRMR.vinh2016can}\\
RJMIM & RJMIM & \cite{RJMIM.peng2016rjmim}\\
SAFE & \makecell{Self-Adaptive Feature Evaluation} & \cite{SAFE.singha2018adaptive}\\
\hline\hline\\  [-2ex]
\end{tabular}
\label{tab:algoritmos}
\end{table*}

\section{Issues and future challenges}
\label{seccion:casos}
%%%%%%%%%%%%%%%%%%%%%%%%%%%%%%%%%%%%%%%%%%
Having seen filter methods that are based on feature intercooperation, some issues arise with maybe subtle distinctions that we'd like to point at, signaling future challenges.

\subsection{Interaction, Synergy, and Complementarity.}
\begin{figure*}[t]
\centering
\includegraphics[width=0.6\linewidth]{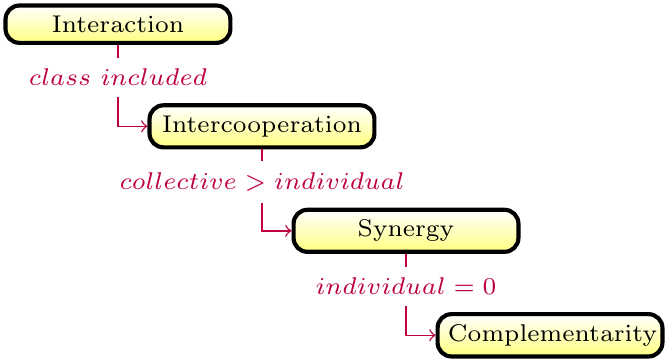}
\caption{Conceptual relationship between terms that differentiates them.}
\label{fig:terminos}
\end{figure*}

In the literature, the terms interaction, synergy, and complementarity are used interchangeably; however, we consider they are not synonymous and have different meanings (Figure \ref{fig:terminos}). 
\textcolor{revisor1}{In this sense, in the most recent study \cite{wan2022r2ci} an interesting distinction is essentially made between complementarity and interaction, from the point of view of attribute generation into search space; in which the significance of features is assessed through their relevance to the class, redundancy and complementarity with selected features, and interaction with remaining unselected features.
Thus, this work complements and extends existing research such that the distinction between interaction, synergy and complementarity is made from the point of view of quantifying multivariate dependencies, and the roles of these variables (i.e. features and/or class).}

In essence, \emph{interaction} is a measure of dependence between $2$ or more variables and can therefore be understood as a nonlinear generalization of correlation. This implies that it can be used to capture a two-way dependence as a minimum order (number of features) in which the class can be included or not. 
\textcolor{revisor1}{
\begin{definition}
There exists \textbf{interaction} among variables $X_1, X_2, ..., X_n$ whenever their multivariate symmetrical uncertainty is positive, that is,
$MSU(X_1, X_2,...,X_n) > 0$.
\end{definition}}
Now let's consider the question of using interaction to measure the amount of information provided by $2$ or more attributes together about the class. In this case, we are talking about \emph{intercooperation} (i.e., multi-way interaction among $2$ or more features and the target).
\textcolor{revisor1}{\begin{definition} 
There exists \textbf{intercooperation} among features $F_1,F_2,...,F_n$ about the class $C$ whenever 
$MSU(F_1, F_2,...,F_n, C) - MSU(F_1, F_2,...,F_n) > 0$.
\end{definition}}
In this sense, the \emph{synergy} term means that the intercooperation among features provides more information about the class label as a whole than the sum of the individual contributions. 
\textcolor{revisor1}{\begin{definition} 
There exists \textbf{synergy} among features $F_1,F_2,...,F_n$ about the class $C$ whenever $MSU(F_1, F_2,...,F_n, C) - MSU(F_1, F_2,...,F_n) > SU(F_1, C) + SU(F_2, C) + \cdots + SU(F_n, C)$.
\end{definition}}
Following the definition above, \emph{complementarity} occurs when attributes individually do not appear to contain any information about the class and can only contribute in combination with others.
\textcolor{revisor1}{\begin{definition} 
There exists \textbf{complementarity} among features $F_1,F_2,...,F_n$ about the class $C$ whenever $[MSU(F_1, F_2,...,F_n, C) - MSU(F_1, F_2,...,F_n) > 0] \land SU(F_i, C) = 0, \forall i \in \{ 1, ..., n \}$.
\end{definition}}

\subsection{Filter method categorization with respect to criterion function scope.}
Feature measure or evaluation criterion plays an important role in feature selection, which forms the basis of feature selection \cite{liu2012libro}. 

Given a target variable $\mathcal{C}$ and $\mathcal{F}$ an $n$ dimensional feature set, where ${f}_i \in \mathcal{F}$ is used for representing its elements. Let $J(\mathcal{X})$ be a criterion function that evaluates a feature subset $\mathcal{X}\subset \mathcal{F}$. Then the feature selection can be formulated as the problem of finding an optimal subset of features $\mathcal{S}_\textit{opt}$ for which
\begin{equation}\label{max:fs:prob}
 J(\mathcal{S}_\textit{opt}) = \max_{\mathcal{X}\subset \mathcal{F}} J(\mathcal{X}). 
\end{equation}

Consider $\mathcal{S}$ as the subset of currently selected features and $f_i$ as a candidate feature to be added to or deleted from $\mathcal{S}$. Based on the criterion function scope, filter selection methods may roughly be divided into:
\subsubsection{1st generation filter methods.}
They can only  measure attributes' relevance according to the amount of individual information contained with respect to the class. These methods are the simplest since their criterion function is defined as:
{\small
\begin{equation}\label{fs:first_gen}
 J(f_\textit{i}) = \textit{IndividualRelevance}(f_\textit{i}, \mathcal{C}). 
\end{equation}
}
\subsubsection{2nd generation filter methods.}
While the individual evaluation is incapable of removing redundant features because redundant features are likely to have a similar amount of information, second-generation filter methods can handle feature redundancy with feature relevance through the criterion function:
{\small
\begin{eqnarray}\label{fs:second_gen}
 J(f_\textit{i}) & = & \textit{IndividualRelevance}(f_\textit{i}; \mathcal{C}) \nonumber \\
 & &\qquad \qquad  - \textit{Redundance}(f_\textit{i}; \mathcal{S}). 
\end{eqnarray}
}
\subsubsection{3rd generation filter methods.}
A clear limitation of previous approaches is that they neglect a feature that appears to be irrelevant or weakly relevant to the class individually, but when it is combined with other features, it may highly correlate to the class. A concept that was recently considered is intercooperativeness, in which a set of two or more features cooperate to provide information about the target concept:
\begin{equation}\label{fs:thirth_gen}
\renewcommand{\arraystretch}{1.5}
\begin{array}{@{} r @{} >{{}} c <{{}} @{} l @{} l @{} l @{} }
  J(f_\textit{i}) & =  & \textit{IndividualRelevance}(f_\textit{i}; \mathcal{C})\\
             &    & - \textit{Redundance}(f_\textit{i}; \mathcal{S})\\ 
             &    & + \textit{Intercooperation}(\{f_\textit{i},..,f_\textit{j}\}; \mathcal{S};\mathcal{C}).\\
\end{array}
\end{equation}

\subsection{Cooperativeness and Exclusive Cooperativeness.}
As shown in \cite{yu2004efficient}, finer classification of attribute types might contribute to exploring novel attribute selection strategies, so we propose a conceptual subdivision of attributes into cooperativeness and exclusive cooperativeness according to either independence level or absolute dependence on other attributes in order to provide information. Namely, a cooperative attribute provides information about the class individually and in cooperation with other attributes, while an exclusively cooperative attribute only becomes relevant in the context of others.

\subsection{Simultaneous Evaluation and Evaluation by Phases.}
In \cite{yu2004efficient}, existing approaches to relevance and redundancy were studied. They defined a traditional approach as one that implicitly manages redundancy of attributes with their relevance (i.e., simultaneous evaluation) and proposed another approach in which redundant attributes are explicitly identified for their elimination (i.e., evaluation by phases).

In this regard, when designing a third-generation filter method, we should consider possible cooperation between attributes and the impacts on the scalability and stability resulting from simultaneous evaluation. Thus, scalability is the sensitivity of the computational performance of the feature selection method to data scale, and stability is the sensitivity of feature selection results to training set variations.

%In this regard, when designing a third-generation filter method, we should consider not only possible cooperation between attributes, but also possible impacts on the scalability and stability resulting from  simultaneous evaluation. Thus, scalability is sensitivity of computational performance of feature selection method to data scale and stability is sensitivity of feature selection results to training set variations.

Novel feature selection methods need to be developed, in which the evaluation by phases is considered. This approach, which decouples individual relevance analysis, redundancy analysis, and intercooperation analysis, offers alternatives for search space reduction.

\subsection{Multivariate Dualist Measures.}

In \cite{timme2014synergy}, an interesting perspective was studied: measures that treat all variables equally and measures that treat the class separately from the group of attributes. We refer to the latter ones as multivariate dualist measures. Although this type of measure has been successfully applied to various fields \cite{lizier2018information}, to the best of our knowledge, the use of dualistic multivariate measures for the selection of attributes has not yet been implemented \cite{yu2018multivariate}. Hence, feature selection based on multivariate dualist measures is an interesting possibility.

\subsection{Maximum Intercooperation Order.}
Previous works on real data sets show that the inclusion of high-order dependencies can improve feature selection based on mutual information \cite{RelaxMRMR.vinh2016can}.

However, the number \emph{k} $(k=2,3,4,5,m)$ of interaction terms is generally determined by expert information, amount of data, degree of error, high-dimensionality assumptions, or some technical considerations such as scalability and/or computation time. As the number of candidate interactions increases exponentially with the number of attributes, it is worth investigating high-order interactions to achieve a balance between accuracy and complexity.

\subsection{Intercooperation Over/Under Estimation.}
Although third generation filter approach overcomes some of the drawbacks of previous generations, it has to deal with new issues. Thus, possible overestimation or underestimation should be considered in the quantification of synergistic information as shown in \cite{griffith2014quantifying}. The detection of intercooperativeness itself is a challenge, and therefore, its precise measurement produces a greater challenge.

\subsection{Redundancy and/xor Synergy.}
Many different groups have developed multivariate measures in use today and differ in subtle but significant ways. Thus, a crucial topic related to multivariate information measures is understanding the relationship and meaning of synergy and redundancy. Some authors argue that redundancy and a synergy component can exist simultaneously, whereas others argue that synergy and redundancy are mutually exclusive qualities \cite{timme2014synergy}.

From the viewpoint of feature selection, the distinction between synergy and redundancy is essential; therefore, their effects are still an open question.

\subsubsection{Inter-feature redundancy term and complementarity effects.}
An interpretation of the objective function of known methods as approximations of a target objective function is proposed in \cite{macedo2019theoretical}. 

In the same paper it is verified that a redundancy consisting of the level of association between the candidate attribute and the previously selected attributes is called inter-feature redundancy. Such redundancy is important, for instance, to avoid later problems of collinearity. Furthermore, feature selection methods that include inter-feature and class-relevant redundancy terms take into account the complementarity expressed as the contribution of a candidate feature to the explanation of the class when taken together with already selected features.

\subsubsection{Evaluating interaction from the addition of features.}
Given a selected feature subset $S_j$ consisting of $j$ variables, suppose we increase the number of variables to $k$ achieving subset $S_k$ so that $S_j \subset S_k$.

If $MSU(S_j) < MSU(S_k)$ the addition of variables has caused a \textit{gain in multiple correlation}, and we can say that the added variables $S_k - S_j$ interact positively with $S_j$. In the opposite case, if $MSU(S_j) > MSU(S_k)$ we can say that the added variables $S_k - S_j$ interact negatively with $S_j$.

A proposal of formal definition for interaction in \cite{sgomez2022} is in terms of $k$-way interaction on top of $j$ variables: It is the minimum gain in multiple correlation over all possible choices of $j$-variable subsets $S_j$ within $S_k$. Note that from a combinatorics point of view, there are $C(k,j)$ possible such subsets.

The proposed definition covers general and complex cases, but it also accommodates the already known classical statistics cases of interaction on a numeric response, occurring in multiple regression and analysis of variance.

\subsubsection{\textcolor{revisor2}{Intercooperation via Game Theory.}}

\textcolor{revisor2} {In recent years, other approaches have been investigated to overcome the limitation associated to traditional information-theory-based measures. One of these approaches that have gained popularity is Game Theory (GT).}

\textcolor{revisor2} {In GT, the different scenarios are mathematically assessed so that the success of an individual decision depends on the decision choices of others~\cite{von1947theory}. Azam and You~\cite{azam2011incorporating} propose to use GT in feature selection to deal with high imbalance situations in text categorization. Sun et al.~\cite{sun2012feature} introduce a cooperative game-theory-based framework to identify the power of each feature according to intricate and intrinsic interrelations among features. Afghagh et al.~\cite{afghah2018game} propose a novel information-theoretic predictive modeling technique based on the idea of coalition game theory for feature selection.}

\textcolor{revisor2} {Within GT, the Shapley Value (SV) has been used for feature selection by Chu and Chan~\cite{chu2020feature}. In this work, the SV is decomposed into high-order interaction components to measure the different interaction contributions among features. Bimonte and Senatore~\cite{bimonte2022shapley} use the SV to construct the weighted contribution for each feature to allow the selection of features that have explanatory value.}

\textcolor{revisor2} {Summarizing, GT, in general, and SV, in particular, are useful approaches to identify the cooperation among features.}

%%%%%%%%%%%%%%%%%%%%%%%%%%%%%%%%%%%%%%%%%%
\subsubsection{\textcolor{revisor2}{Feature Selection and/or Deep Learning.}}
\textcolor{revisor2}{Deep Learning (DL) \cite{lecun2015deep} is an advanced sub-field of Machine Learning that simplifies the modeling of various complex concepts and relationships using multiple levels of representation. DL is distinct from feature selection as DL leverages deep neural networks structures to learn new feature representations while feature selection directly finds relevant features from the original features, thus yielding more readable and interpretable results \cite{li2017feature}}. 

\textcolor{revisor2}{Although DL techniques for attribute selection have shown good results, we believe that more attention should be paid to the importance of attributes and the interpretability of machine learning models, since the most accurate estimates are not always sufficient to solve a data problem.} 

\textcolor{revisor2}{On the other hand, several studies show that the use of attributes filtered by traditional attribute selection methods and their use as input in a deep generative model outperforms state-of-the-art approaches. Therefore, the study of the effects of intercooperation-based attribute selection methods in a deep generative predictive model are still an open question.}

%%%%%%%%%%%%%%%%%%% RESUMEN DE LA SECCION %%%%%%%%%%%%%%%%%%%%%%%
%\textcolor{revisor1}{We have seen that detection and measurement of intercooperation employs a variety of techniques. It seems advisable to evaluate the quality of results for each one of them after appropriate criteria were set.}

%\textcolor{magenta}{NO ENTENDI EL PARRAFO DE ARRIBA}
%%%%%%%%%%%%%%%%%%%%%%%%%%%%%%%%%%%%%%%%%%
\section{Conclusions}
%%%%%%%%%%%%%%%%%%%%%%%%%%%%%%%%%%%%%%%%%%
Feature selection plays an important role in knowledge discovery. It is an effective technique in dealing with dimensionality reduction, removing irrelevant data, increasing learning accuracy, and improving result comprehensibility. Therefore, feature selection is active research in the fields of data mining and machine learning. Over the past decade, most research in filter methods has emphasized the use of feature intercooperation to assist in the feature subset selection process. In this paper, we have surveyed $27$ filter feature selection methods that adopt this approach, covering important gaps in the field. In addition, the concepts of relevance, redundancy and intercooperativeness are defined and quantified through information theory measures. Finally, the most significant issues and challenges of filter methods based on feature intercooperation are described, identifying the future research directions in this area.

\backmatter

\bibliography{sn-bibliography}% common bib file


\end{document}